\documentclass[11pt]{article}
\usepackage{latexsym,amsmath,amssymb,amsthm,enumerate,color,graphicx,psfrag,textcomp}
\usepackage{amssymb}
\usepackage{geometry}
\newtheorem*{Rule*}{Deletion Rule}

\usepackage{algorithm, algorithmic}
\usepackage{caption}
\usepackage{subfigure} 
\usepackage{svg}
\usepackage{diagbox}
\usepackage{multirow}
\usepackage{booktabs}

\usepackage{setspace}
\usepackage{rotating} 
\usepackage[colorlinks=false]{hyperref}
\usepackage{natbib}
\usepackage{wrapfig}
\usepackage{enumitem}

\begin{document}
\begin{center}
{\Large \bf{Enhancing Column Generation by Reinforcement Learning-Based 
\vskip 0.2cm
Hyper-Heuristic for Vehicle Routing and Scheduling Problems}}

\vskip 0.2cm
{\bf Kuan Xu$^{a}$, Li Shen$^{b}$, Lindong Liu$^{c}$\footnote{Corresponding author.}} \\ 
\vskip 0.2cm

{\small $^a$International Institute of Finance, School of Management, University of Science and Technology of China, 230026, P.R. China,  mathkxu@mail.ustc.edu.cn}\\
{\small $^b$JD EXPLORE ACADEMY, 100000, P.R. China,  mathshenli@gmail.com}\\
{\small $^c$International Institute of Finance, School of Management, University of Science and Technology of China, 230026, P.R. China,  ldliu@ustc.edu.cn}\\
\end{center}

% Here goes the abstract
\begin{quote}
{\bf Abstract:}
Column generation (CG) is a vital method to solve large-scale problems by dynamically generating variables. It has extensive applications in common combinatorial optimization, such as vehicle routing and scheduling problems, where each iteration step requires solving an NP-hard constrained shortest path problem.
Although some heuristic methods for acceleration already exist, they are not versatile enough to solve different problems. In this work, we propose a reinforcement learning-based hyper-heuristic framework, dubbed RLHH, to enhance the performance of CG. RLHH is a selection module embedded in CG to accelerate convergence and get better integer solutions. In each CG iteration, the RL agent selects a low-level heuristic to construct a reduced network only containing the edges with a greater chance of being part of the optimal solution. In addition, we specify RLHH to solve two typical combinatorial optimization problems: Vehicle Routing Problem with Time Windows (VRPTW) and Bus Driver Scheduling Problem (BDSP). The total cost can be reduced by up to 27.9\% in VRPTW and 15.4\% in BDSP compared to the best lower-level heuristic in our tested scenarios, within equivalent or even less computational time. The proposed RLHH is the first RL-based CG method that outperforms traditional approaches in terms of solution quality, which can promote the application of CG in combinatorial optimization.

{\bf Key words:} Column Generation; Reinforcement Learning; Hyper-Heuriistic; Vehicle Routing Problems; Crew Scheduling Problems 
\end{quote}

\normalsize

\section{Introduction}

% 介绍 CG in routing and crew schdueling
Column generation (CG), also referred to as branch-and-price when integrated into a branch-and-bound framework, is first applied to cutting stock problems and had a great impact in the field of OR \citep{gilmore1961linear,gilmore1963linear}. It solves problems with exponential large solution space by dynamically generating variables and has proven to be highly effective in tackling a diverse range of combinatorial optimization (CO) problems \citep{ColumnGeneration2005}. 
It decomposes the primary problem into two parts: a restricted master problem (RMP) that is consistent with the original linear program but limited to a subset of variables, and one or more pricing problems (PP) to generate new columns (representing variables) that enhance the current RMP solution. 
In most vehicle routing or crew scheduling applications, the problems are defined on a network. The master problem (MP) is a set partitioning or set covering problem with side constraints, where the variables are associated with vehicle routes or crew schedules. These variables, denoted as columns in CG, are generated by PP that corresponds to a shortest path problem with resource constraints (SPPRC) or elementary SPPRC (ESPPRC). An elementary path is a path in which all nodes are pairwise distinct.

% 说明为什么要使用启发式方法约简定价网络
% 本质上的难点，要有宏观（方法层面）的 comment
It's worth noting that the SPPRC is commonly known to be NP-hard \citep{dror1994cg4vrp} and needs to be solved in every iteration of CG. According to \cite{ejor2018accelerating}, the solution of PP consumes more than 90\% of the CPU time in the CG process, so it is almost impossible to solve vehicle routing and scheduling problems as the problem size grows large.
Researchers have been using various heuristic methods to speed up the PP by reducing the size of the pricing network \citep{bestedges1,bestedges2,bestedges3}.
% Yet, different heuristics may be applicable to different problems or even different iterations of the same problem instance. There is (as far as we know) no good approach for how to pick an appropriate heuristic for each iteration.
And \cite{morabit2022} propose a ML-based pricing heuristic and show that it can speed up the convergence of the CG process.
But their ML model does not perform as well as a simple heuristic algorithm for VRPTW.
So how to make full use of the advantages of different heuristic methods is worth studying.

In this work, we propose a hyper-heuristic method based on deep reinforcement learning, dubbed RLHH, embedded in the iterative framework of CG to enhance its performance for various vehicle routing and crew scheduling problems. Our method combines the powerful information extraction and generalization capabilities of neural networks with the theoretical advantages of the CG algorithm.
Specifically, the method takes advantage of the solution information of RMP and features of pricing networks in PP to select the appropriate low-level heuristic algorithm, constructing a reduced network containing the edges that have a greater chance of being part of the optimal solution.
The reduced network is used at each CG iteration if it generates a satisfactory column. Otherwise, we use the complete network, especially in the last iteration.
In the RL context, we use a Double Deep Q Network algorithm (DDQN) with experience replay \cite{nature2015dqn} to train a Q-function approximator which determines the optimal decision for the agent in each state.

We apply our proposed RLHH on two well-known CO problems: the vehicle routing problem with time windows (VRPTW) \citep{VRPTW1992,VRPTW1999} and the bus driver scheduling problem (BDSP) \citep{smith1988bdsp,bdsp2022column}, which have different network structures. Still, our RLHH method improves the CG performance by pruning pricing networks, making the method presented in this paper general enough for different problems. 
In the VRPTW experiments, RLHH achieves a 27.9\% reduction in total cost within 75\% of the computation time compared to the best low-level heuristic method. Similarly, in BDSP, the reductions in the total cost of up to 15.4\% can be obtained within a comparable computational time. 
Moreover, our experiments illustrate that RLHH has a strong generalization capability to more challenging problems, in terms of problem structure and problem scale.  

In the end, we summarize our main contributions as below:
\begin{itemize}[itemsep=5pt,parsep=0pt]
    \item We propose a reinforcement learning-based hyper-heuristic (RLHH) method to enhance the column generation (CG) for combinatorial optimization (CO) problems. It shows the great potential of combining the stability and efficiency of OR algorithms with learning and generalization capabilities of RL models.
    \item We specify RLHH to solve two classic CO problems: VRPTW and BDSP. Compared with previous work, we consider the quality of the final solution in the reward function. The experimental results demonstrate that our method can achieve better solutions than a single heuristic method within the same or even shorter time frame.
    \item We conduct extensive experiments to show that RLHH has a strong generalization for problems with more complex structures or larger scales, which are rarely studied previously, and outperforms traditional methods. 
\end{itemize}

% Literature Review
\section{Literature Review}
% 补充CG应用，强调CG的重要性
% \textcolor{red}{Reinforcement Learning has found applications in diverse domains, including robotics, game playing, recommendation systems, and autonomous systems. It excels in scenarios where explicit programming is infeasible, and systems must learn from experience to make informed decisions in complex and dynamic environments.}

% ML4CO
% \paragraph{Learning for Combinatorial Optimization.}
Machine Learning (ML) for combinatorial optimization has recently received much attention \citep{bengio2021ML4CO,lombardi2018}. \cite{yan2020} describe how learning techniques and paradigms are applied to CO, specifically Graph Matching.
Routing problems, especially TSP and VRP, have been explored by a sequence of works \citep{vinyals2015pointer,khalil2017learning,kool2018attention}. Most of these works follow an end-to-end approach, which is directly constructing solutions by ML models.

We focus on a literature series that exploits ML to assist optimization problem-solving.
Many studies apply ML to branch-and-bound solvers, such as learning variable selection method \citep{khalil2016learn2branch,gasse2019learn2branch,branch2022lookback} and learning node selection \citep{furian2021machine}. \cite{ijcai2017Learn2RunHeuristics} use ML models to dynamically decide whether to run a heuristic at each node of the search tree. 
And many researchers use ML to assist or control the local search \citep{ma2021learning,hottung2022neural,ijcai2022NNS4PDP}, among which \cite{lu2020learn2improve} present the first RL framework that outperforms classical Operations Research (OR) solvers LKH3 on capacitated VRP. And \cite{jacobs2021} solve robust optimization utilizing an RL-based heuristic method, \cite{ejor2022rlhh} present an RL-based hyper-heuristic for CO problems with uncertains.

% ML+CG, 说明我们的方法与现有工作的区别
% \paragraph{Column Generation.}
CG is an indispensable tool in CO to solve large-scale integer programming. 
\cite{morabit2021} exploit supervised learning to select promising columns at each iteration in the CG procedure. The experiments show that their approach accelerates the CG algorithm by 30\%. However, it requires labeled data generated by an extremely time-consuming mixed-integer linear program. 
\cite{RLCG2022} use the RL model that avoids this problem, and the proposed selection module converged faster than the greedy column selection strategy. 
\cite{babaki2022coil} propose a neural CG architecture to select columns in each iteration by imitation learning. Both \cite{morabit2021} and \cite{RLCG2022} use ML for column selection to speed up the CG procedure, while our method greatly accelerates the pricing problem, which usually consumes most of the computing time.

The most related to our proposed RLHH is \cite{morabit2022}, which proposes a new heuristic pricing algorithm based on ML to prune the pricing network and accelerate the PP. Similarly, \cite{yuan2022cgp} present a graph neural network-based pricing heuristic for railway crew scheduling problems (RCSP). They utilize supervised learning, which requires a large amount of labeled training data to be obtained in advance through the OR solver.
Even so, the trained ML model does not perform as well as a heuristic method in VRPTW experiments \citep{morabit2022}. And \cite{qin2021novel} develops a reinforcement learning-based hyper-heuristic to solve the heterogeneous vehicle routing problem, which outperforms the existing meta-heuristic algorithms.
We exploit different simple and effective low-level heuristic algorithms to construct an RL-based hyper-heuristic model. The training data is generated during the agent and environment interactions in the RL framework. As a result, we can circumvent the issue of acquiring hard-to-obtain training data or dealing with low-quality labels often encountered in supervised learning.

\section{Preliminaries}
In this section, we initially present the fundamental concepts of CG along with the essential notation. Subsequently, we delve into the pricing problem that is the primary focus of this article: SPPRC. Finally, we provide a concise summary of RL terminology to enhance the comprehension of the RLHH method proposed in this paper.

\vspace{-0.1cm}
\subsection{Column Generation}
\label{section:CG}

Let us focus on the linear relaxation of the original integer programming, considering the following master problem (MP):
\begin{align}
    z^*_{MP} := & \min_x~~ \sum_{p\in\Omega} c_p x_p \\
    \text{s.t.}~~ & \sum_{p\in\Omega} \boldsymbol{a}_{\boldsymbol{p}} x_p = \boldsymbol{b}, \label{eq:constraints} \\ 
    & \quad\quad\quad~ x_p \ge 0, ~~ \forall{p\in\Omega},
\end{align}
where $\Omega$ is the set of variable indices, i.e., the set of feasible routes or schedules; 
$c_p\in\mathbb{R}$ represents the variable cost, $\boldsymbol{a}_{\boldsymbol{p}} \in \mathbb{R}^m$ the constraints coefficients and $\boldsymbol{b} \in \mathbb{R}^m$ the constraints right-hand side vector.
When $|\Omega|$ is prohibitively large so that the variables (columns) cannot be enumerated explicitly, we consider a restricted master problem (RMP) that only contains a subset $\mathcal{F} \subseteq \Omega$.

We optimize RMP at each CG iteration and get the optimal dual solution $\pi\in\mathbb{R}^m$ with respect to constraints (\ref{eq:constraints}). Then, the pricing problem (PP) is established using the dual values $\pi$ to search for new columns with the least reduced cost:
\begin{flalign}
    &&
    \min_{p\in\Omega}~~ c_p - \pi^T\boldsymbol{a}_{\boldsymbol{p}},
    &&
\end{flalign}
which is an ESPPRC in the VRPTW case or SPPRC in the BDSP case. If the new column has a negative reduced cost, it's added to the RMP to improve its solution further. Otherwise, RMP has captured all the necessary columns, which means the optimal solution of MP is obtained. 
\textcolor{blue}{
Finally, by limiting the variables of RMP to integers and solving the integer RMP (IRMP), we obtain a integer solution as the result of the experiment in this paper.
}

% \vspace{-0.1cm}
\subsection{SPPRC Formulation}
\label{appendix:SPPRC}

The standard SPPRC often appears in problems defined on acyclic networks, such as vehicle and crew scheduling problems of which BDSP is a special case \citep{Haase2001vcsp} and aircrew rostering problem \citep{rostering1999}. Another well-known variant is the elementary SPPRC (ESPPRC) occurring in vehicle routing problems (an elementary path is a path in which all nodes are pairwise different) \citep{vrp2019}, which is more challenging.

Let $G=(V,E)$ be a directed graph with a set of nodes $V$ that include the source and destination nodes denoted by $s$ and $t$ respectively, $E$ the set of edges. And let $R$ the set of resources. 
% We continue to use the notation for RMP in Section~\ref{section:CG}: $c_p\in\mathbb{R}$ represents the variable cost, $\boldsymbol{a}_{\boldsymbol{p}} \in \mathbb{R}^m$ the coefficients of the constraints, $\boldsymbol{b} \in \mathbb{R}^m$ the constraints right-hand side vector and the dual values $\pi$.
Notably, the number of edges in the pricing network $|E|$ is proportional to $n^2$, where $n$ is the number of customers in VRPTW or trips in BDSP. In a full CG process, it may be necessary to solve PP hundreds of times. It is highly non-trivial to derive a suitable reduced network to solve PP more efficiently.
Without acceleration techniques, it is almost impossible to obtain quality solutions in an acceptable computation time \citep{ColumnGeneration2005}. 

The PP aims to find the optimal column (i.e. route or schedule) on network G that satisfies the resource constraint, so we call G the pricing network.
For each node $i\in V$, let $T_i^r$ represents the value of resource $r\in R$ accumulated on a patial path from node $s$ to $i$, constrained by the resource windows $[\underline{w}_i^r, \overline{w}_i^r]$. 
% where $\underline{w}_i^r, \overline{w}_i^r \in \mathbb{R}$ limiting the values that each resource $r\in R$ can take on node $i$. 
For each edge $(i,j)\in E$, we define the cost $c_{ij}$ and the consumption for each resource $t_{ij}^r\in\mathbb{R}, r\in R$.
Let $X_{ij}$ be the decision variables of the model, which represent the flow on the edges $(i,j)\in E$. 
The PP for each iteration is then formulated as the following SPPRC:
\begin{align}
    &\label{obj} \min~~ \sum_{(i,j)\in E} c_{ij}X_{ij} \\
    \text{s.t.}~~ 
    &\label{flow_s} \sum_{j\in V} X_{sj} - \sum_{j\in V} X_{js} = 1, \\
    &\label{flow_i} \sum_{j\in V} X_{ij} - \sum_{j\in V} X_{ji} = 0, ~~ \forall i\in V\backslash \{s,t\}, \\
    &\label{flow_t} \sum_{j\in V} X_{tj} - \sum_{j\in V} X_{jt} = -1, \\
    &\label{res_consume} X_{ij}(T_i^r + t_{ij}^r) - T_j^r \le 0, ~~\forall(i,j)\in E, \\
    &\label{res_bound} \underline{w}_i^r \le T_i^r \le \overline{w}_i^r,~~ \forall r\in R, \forall i\in V, \\
    &\label{binary} X_{ij} \in \{0,1\},~~ \forall(i,j)\in E,
\end{align}
where the objective~\eqref{obj} represents the total cost of the path, constraints~\eqref{flow_s}-\eqref{flow_t} ensure the flow conservation along the path, while constraints~\eqref{res_consume} model the resource consumption for each resource $r\in R$ on edge $(i,j)\in E$. 
Constraint~\eqref{res_bound} restricts the resource values accumulated in the resource intervals at each node. The last constraints~\eqref{binary} are the binary requirements on the decision variables $X_{ij}$.
% Note that there are more complex SPPRC than those expressed by~\eqref{obj}-\eqref{binary}, i.e., using complex resource extension function that, given the resource values $T_i^r$ at node $i$, provides a lower bound on the resource values at node $j$.

The dual values $\pi$ are used to find new paths (i.e., columns in CG terminology) in a pricing network at each CG iteration. Consider a new path $p\in\Omega$ covering a set of edges denoted as $E_p$ and servicing a set of customers denoted as $V_p$. Then if we add a path $p$, the amount of change that will bring to the total cost (i.e., the reduced cost) is:
\begin{align}
    \Delta_p = \sum_{(i,j)\in E_p} c_{ij} - \sum_{v\in V_p}\pi_v.
\end{align}
We define a binary parameter $\rho_{ij}^k$, taking value 1 if the edge $(i, j) \in E$ contributes to the master problem constraint $k$, 0 otherwise. Then the reduced cost $\Delta_p$ can be rewritten as follows:
\begin{align}
    \Delta_p = \sum_{(i,j)\in E_p} c_{ij} - \sum_{(i,j)\in E_p} \sum_{k\in V_p}\rho_{ij}^k\pi_k
    = \sum_{(i,j)\in E_p} \Big(c_{ij} - \sum_{k\in V_p}\rho_{ij}^k\pi_k \Big).
\end{align}
For each edge $(i,j)\in E$, we define the modified cost on it as $\overline{c}_{ij} = c_{ij} - \sum_{k\in V} \rho_{ij}^k\pi_k$. 

Therefore, solving problem \eqref{obj}-\eqref{binary} by using the modified costs in~\eqref{obj} yields a column with the least reduced cost. 
The SPPRC and ESPPRC are typically solved by dynamic programming based approaches. We employ a labeling algorithm proposed in \cite{ejor2017labeling} in our experiments. We would like to mention that our RLHH framework is applicable regardless of the specific algorithm used for PP.

\subsection{RL Terminology}
Reinforcement learning is a branch of machine learning concerned with the study of sequential decision-making tasks. It frames the learning problem as an agent interacting with an environment over discrete time steps. At each time step $t$, the agent observes the representation of the current environment state $S_t\in S$, selects an action $A_t\in A(S_t)$, and receives a scalar reward $R_{t+1}\in R$ from the environment. The objective of the agent is to learn a policy $\pi_t(a|s)=Pr\{A_t=a|S_t=s\}$ that maximizes the expected cumulative reward over time. A transition is denoted by $(s,a,r,s')$, where $s,a$ and $r$ are the state, action and reward of the current step respectively, and $s'$ is the next state after the agent executes the action $a$. We use an experience replay buffer $M$ to store the transition data generated during the interaction. RL encompasses exploration strategies to balance the trade-off between gathering information about the environment (exploration) and exploiting known knowledge to maximize rewards (exploitation).

% In this paper, we view the CG solution framework as the environment of RL. 
% In Reinforcement learning (RL), the learner (i.e., agent) interacts with the environment (i.e., the CG solution procedure in this paper) and tries to maximize the rewards of the decisions (i.e., actions) it makes. In this context, the agent learns by trial and error while optimizing its reward function. After a number of iterations, a Q-function is learned to tell the agent what is the best decision to make in each situation (i.e., state).

\section{Methodology}
Our goal is to achieve a reduced SPPRC network by selecting the best heuristic method to retain the most promising arcs. This significantly enhances the efficiency of finding new columns in each step of the CG iteration process. It is a sequential decision task, which can be modeled as Markov decision process and solved by RL algorithm.

\subsection{Low-level Heuristics}
\label{appendix:baseline}
Due to the difficulty of the problem, many SPPRC heuristics have been proposed in the literature. \cite{bestedges3} relax the dominance rule by only dominating on a subset of the resources when solving with a labeling algorithm. And several strategies are based on reducing the size of the pricing network, which is the focus of this article. Below we describe in detail the five low-level algorithms we use.

\paragraph{BestEdges1 (BE1).}
\cite{bestedges1} utilize a CG-based approach to solve the Vehicle Routing Problem with Simultaneous Distribution and Collection, which is the variation of the capacitated vehicle routing problem. They remove edges that satisfy:
\begin{equation}
    c_{ij}>\alpha_1 \pi_{max}, ~ \text{where} ~ \pi_{max} = \max_k\{\pi_k\},
\end{equation}
where $\alpha_1\in(0,1)$ is a parameter and $\pi$ is the dual values. This approach may be more suitable for the relatively significant cost difference between edges. And the parameter $\alpha_1$ is to find a trade-off between the quality of the solution and the speed of solving the PP.

\paragraph{BestEdges2 (BE2).}
\cite{bestedges2} solve the problem of designing a container liner shipping feeder network using a CG-based algorithm. In the PP, they sort the edges by their reduced cost $\overline{c}_{ij}$ and only keep the $\alpha_2|E|$ ones with the lowest value, where $\alpha_2\in(0,1)$ is a parameter (ratio). Different from BE1, this method considers the modified edge cost, which contains the original cost and dual information of the RMP. So in a sense, BE2 uses more information than BE1, and we also find that the effect of BE2 is indeed better in our experiments (see Section~\ref{section:experiments} for details).

\paragraph{BestEdges3 (BE3).}
\cite{bestedges3} exploit a CG-based method to solve the VRPTW. In order to accelerate the solution of PP, they proposed a strategy to prune the pricing network: sort the incoming and outgoing edges of each customer node by their reduced cost $\overline{c}_{ij}$, where only $N$ in-edges and $N$ out-edges with the least reduced costs are kept. The most significant difference between this approach and the previous two is that it takes into account all customer nodes, ensuring that each customer has the potential to be served in the PP.

\paragraph{BestNodes (BN).}
In addition to BE2, \cite{bestedges2} suggest another way to obtain a reduced network based on the dual values $\pi$. Firstly, normalize $\pi$ into the interval $[0,\beta]$ according to the formula~\eqref{bn_norm}, where $\beta$ is a parameter. 
\begin{equation}
    \pi^{\prime}_k=\beta\cdot\frac{\pi_k - \pi_{min}}{\pi_{max}-\pi_{min}}, \quad\beta\in(0,1).
    \label{bn_norm}
\end{equation}
Let $\overline{i}$ and $\underline{i}$ the costomers corresponding, respectively, to $\pi_{max}$ and $\pi_{min}$. Then edge $(i,j)$ is removed with a probability directly proportional to the dual value of its target node, with edges incoming to $\underline{i}$ having probability $0$, and edges incoming to $\overline{i}$ having probability $\beta$. This sparsification method only considers the dual values from the RMP, independently from the original cost of edges. It seems effective in cases where the edge cost of the original network is uniformly distributed, and the dual value plays a decisive role.

\paragraph{BestPaths (BP).}
There is a heuristic algorithm based on the shortest path problem without negative cost that is mentioned in \cite{ColumnGeneration2005}. It needs to normalize $\overline{c}$ into the interval [-1,1] first and then set all negative ones to 0, as shown in formula~\eqref{bp_norm}. 
\begin{equation}
    \overline{c}^{\prime}_{ij}=\max\left(0, \frac{(\overline{c}_{ij} - \overline{c}_{min}) - (\overline{c}_{max} - \overline{c}_{ij})}{\overline{c}_{max}-\overline{c}_{min}}\right),
    \label{bp_norm}
\end{equation}
where $\overline{c}_{max}=\max_{(i,j)\in E} \{\overline{c}_{ij}\}$ and $\overline{c}_{min}=\min_{(i,j)\in E} \{\overline{c}_{ij}\}$ represent the maximum and minimum edge modified costs in the pricing network respectively. Finally, compute $K$ shortest paths ($K$ is a parameter) on the normalized network and obtain a reduced network that only contains the edges that belong to these paths. Obviously, this method is more time-consuming than the other four, as Figure~\ref{fig:iteration_process_vrptw} confirms.

\paragraph{}
Last but not least, the parameters in these low-level heuristics are time-evolving. In order to consider solving quality and speed, we try different values of parameters successively. In all experiments for VRPTW and BDSP, we took parameter values according to Table~\ref{tab:baseline_parameter_values}. If all parameter values are tried and no column with a negative reduction cost is found, the heuristic fails and we use the compelet network.

\begin{table}[!htbp]
\setstretch{1.5}
    \centering
    \caption{The parameter values we used in the experiment}
    \label{tab:baseline_parameter_values}
    % \vspace{-0.2cm}
    \begin{tabular}{ccc}
        \toprule
        Name & Parameter & Values \\
        \midrule
        BE1 & $\alpha_1$ & [0.1, 0.3, 0.5, 0.7] \\ 
        BE2 & $\alpha_2$ & [0.1, 0.2, 0.3] \\
        BE3 & $N$        & [0.3, 0.5, 0.7] $*|V|$ \\
        BN  & $\beta$    & [0.9, 0.7, 0.3] \\
        BP  & $K$        & [3, 5, 7, 9] \\
        \bottomrule
    \end{tabular}
\end{table}

\subsection{RLHH Framework}
\label{section:RLHH_framework}
\begin{figure*}[htbp]
    \centering
    \includegraphics[width=0.95\textwidth]{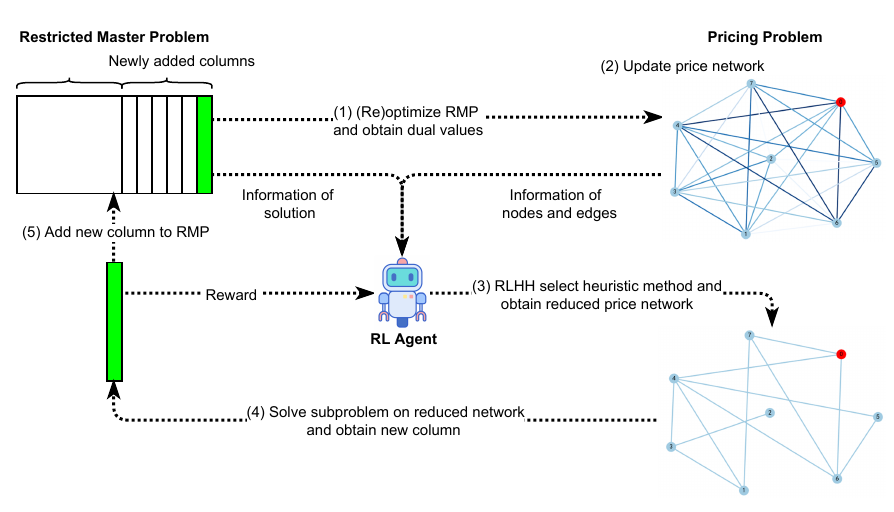}
    \vspace{-0.4cm}
    \caption{Our RL-based hyper-heuristic framework in column generation}
    \label{fig:RLHH_framework}
     \vspace{-0.2cm}
\end{figure*}

The RL-based hyper-heuristic (RLHH) framework in CG (see Figure~\ref{fig:RLHH_framework}) combines the strength of heuristics, which are powerful since they are custom-made for vehicle routing and scheduling problems, with the learning and generalization capabilities of RL.
For the RL model, the CG process as the environment and low-level heuristics constitute our action space. These heuristics generate reduced networks based on their rules and are computationally light. The state includes solution information of RMP and features of the pricing network in PP. With the current state as input, a neural network produces the action score vector, and the weights of the neural network are trained by the DDQN algorithm with experience replay \citep{nature2015dqn}. 
As shown in Section~\ref{section:experiments}, whether it is VRPTW or BDSP, our RL model can distinguish the efficient low-level heuristic in each CG iteration step to generate new valid columns that ultimately leads to excellent integer solutions.

\paragraph{State Space.}
In RLHH, the environment is the CG solution process corresponding to a given problem instance. 
Each state includes features from the current relaxation solution and individual pricing network. 
% These features are described in the Appendix~\ref{appendix:RLHH}.
Fearures used for VRPTW and BDSP considered in this paper:
\begin{itemize}
    \item Current objective value / initial objective value
    \item Sum of fractional solutions / number of fractional solutions
    \item Number of fractional solutions / number of variables 
    \item Statistics and information entropy for dual values, cost and reduced cost of edges
    \item For each resource, the Coefficient of Variation of the resource consumption along edges 
    % \footnote{Coefficient of Variation is defined as the ratio of the standard deviation to the absolute value of the mean.}
\end{itemize}

\paragraph{Action Space.}
The agent selects an action $a_t\in\{BE1, BE2, BE3, BN, BP\}$ at each CG iteration, and then we compute the reduced network using the corresponding heuristic. 
The parameters in these heuristic methods are time-evolving. 
Taking BE1 as an example, we observe that the algorithm finds a lot of negative reduced cost columns, including the optimal one at the first iteration, with $\alpha_1 = 0.1$.
However, if the value of $\alpha_1$ is left unchanged, the algorithm fails to find negative reduced cost columns in the subsequent runs.
Therefore, we sequentially increase the value in the subsequent trials. If no column with negative reduced cost is found with $\alpha_1=0.7$, the heuristic method fails, and we use the complete network. 
% A detailed description of these low-level heuristics can be found in Appendix~\ref{appendix:baseline}.

\paragraph{Reward Function.}
In previous work, \cite{RLCG2022} use an advantage-based reward function $(\text{obj}_{t-1} - \text{obj}_t) / \text{obj}_0$, where $\text{obj}_{t}$ is the objective value of the RMP$_t$. It focuses on the quality of the solution improvement at each iteration, i.e., the change in the RMP objective value. We observe that actions often achieve significant cost reductions in early iterations while doing so becomes increasingly difficult in later steps. In particular, the potential for objective value reduction, and the magnitude of that reduction, decreases with each iteration. Therefore, it seems unfair to give larger rewards for early actions.

In the CG solution framework, we encourage the RL agent to choose the action that can generate a reduced network containing negative reduced cost columns. So we set the step reward $r_t$ as follows:
$$
r_t = \left\{
\begin{aligned}
1& \quad \text{if agent find a column and improve the RMP solution} \\
0& \quad \text{if agent find a column but not improve the RMP solution} \\
-1& \quad \text{if agent can not find a new column in the reduced network}
\end{aligned}
\right.
$$
And more importantly, we want to end up with a good solution. Therefore, at the end of an episode, we use integrality gap to measure the quality of the integer solution:
$$
    r_T = 100^{GAP}, \text{where}\ GAP = \frac{\text{obj}_{int}}{\text{obj}_{frac}},\nonumber
$$
where $\text{obj}_{int}$ and $\text{obj}_{frac}$ represent the objective value of the integer solution and the relaxed solution respectively.
Compared with previous work, we not only consider the performance of the agent in the intermediate steps but also consider the final solution quality. Therefore, their approach merely accelerates CG, whereas our method not only speeds up the process but also attains higher-quality solutions.

\begin{algorithm}[h]
\setstretch{1.5}
    \vspace{0.2cm}
    \caption{RLHH: RL-based hyper-heuristic in Column Generation}
    \vspace{0.2cm}
    \label{alg:RLHH}
    \textbf{Input}: Problem instance \& trained Q-function \\
    \textbf{Output}: Integer solution of the original CO problem\\
    \vspace{-0.6cm}
    \begin{algorithmic}[1] %[1] enables line numbers
        \STATE Let $t=0$; Build the RMP$_0$ with initial columns
        \STATE Solve RMP$_0$; Use dual values to get the complete pricing network. 
        \STATE Calculate initial state $s_0$.
        \WHILE{exist new columns with negative reduced cost}
        \STATE $a_t^*=\arg\max_{a_t\in A_t} Q(s_t, a_t)$; \\
        Utilize selected low-level heuristic to get a reduced network.
        \STATE Attempt to find a new negative reduced cost column in the reduced network. 
        \IF {not found}
        \STATE Find a new negative reduced cost column in the complete network. \label{full_network}
        % \ELSE
        % \STATE Perform task B.
        \ENDIF
        \STATE Add new column to RMP$_t$ and solve RMP$_{t+1}$;
        \STATE Solve RMP$_t$; Use dual values to get the complete pricing network.
        \STATE Calculate next state $s_{t+1}$.
        \STATE Set $t = t+1$
        \ENDWHILE
        \STATE Set the variable to integer type and solve.
        \STATE \textbf{return} Final IRMP solution
    \end{algorithmic}
\end{algorithm}

\paragraph{Training and Execution.}
The RL agent is represented as a value function $Q(s, a)$ related to state $s$ and action $a$, and corresponds to the heuristic selection module in the CG iteration for PP. At each iteration, the agent selects an action according to the state and acquires reward feedback from the PP solution. The Q function is defined as the expectation of discounted cumulative rewards, as denoted in Equation~\eqref{Q-function}, where $\gamma$ is the discounted factor, $t$ is the time step, $R$ is the reward and $\pi$ is the policy. The details of the training process are shown in Algorithm~\ref{alg:RLHH_train} of Appendix~\ref{appendix:RLHH}.
\begin{flalign}
    \label{Q-function}
    &&
    Q^{\pi}(s,a) = E[R_{t}+\gamma R_{t+1}+\gamma^2 R_{t+2}+\cdots | S_t=s,A_t=a,\pi].
    &&
\end{flalign}

Algorithm~\ref{alg:RLHH} shows how a trained RLHH agent is applied in CG to solve CO problems.
The initialization steps 1-2 add an initial set of columns (for instance, in VRPTW, we start with a separate vehicle for each customer) into the basis and build the first pricing network from dual values. 
In the while loop, the RL agent selects the best action from the candidate low-level heuristic method and gets a reduced network.
Then we attempt to find a new column in the reduced network with a negative reduced cost, which could theoretically reduce the value of the objective function.
If that fails, we go to Step~\ref{full_network} to find such a new column on the complete network. The RMP is updated in the same way as the traditional CG method. Finally, the variable is set to the integer type, and the original integer programming is solved.

\section{Experiments}
\label{section:experiments}
% 提出的框架一般可以适用于哪些问题，为什么我们选择下面的问题做实验？
The proposed framework can be applied to various routing problems and scheduling problems. In this paper, we choose two classic cases for experiments, namely VRPTW and BDSP.
In recent years, VRPTW is increasingly becoming an invaluable tool for modelling all aspects of supply chain design and operation. BDSP is a important problem in urban mass transit. As long as appropriate modelling methods are adopted, our method can be directly applied to other scheduling problems, such as RCSP.
Routing problems (such as VRPTW) and scheduling problems (such as BDSP) have different network structures. Still, our RLHH method improves CG performance by selecting promising edges, making the methods described in this paper general enough to be applied to different problems.

We follow \cite{morabit2021} by focusing on solving the root node of the branch-and-price (B\&P) search tree, while the proposed RLHH strategy is also applicable to other nodes. We compare the five low-level heuristics described in section~\ref{section:RLHH_framework} and report two metrics (bound and running time) to show that our method can obtain better integer solutions (upper bounds) in less time. During the B\&P process, a node can be safely pruned without further expansion if its lower bound is no better than the global upper bound, thus significantly reducing the search time for the optimal solution. 

We use a single RTX 3060 GPU for RL training, implement with Python 3.7 and PyTorch 1.10.0. And we use Gurobi 9.1.2 to solve the final IRMP. All experiments implemented on a personal computer with an i7-11700 CPU at 2.5GHz and 16 GB RAM. A full training session consists of 10,000 episodes and takes about 50 hours.

\subsection{Vehicle Routing Problem with Time Windows}
% VRPTW 结果路径
\label{section:VRPTW}

VRPTW is an extension of the capacitated vehicle routing problems, with the vehicle must begin service within the customer's time window. It aims to find feasible routes (each route requires a vehicle) to deliver goods to a geographically dispersed group of customers while minimizing total travel costs. A feasible route is issued from the depot, going to the depot, and satisfying capacity and time window constraints. The most efficient methods to solve this problem are based on CG. We can model the problem as MP in Section~\ref{section:CG}, and each column corresponds to a feasible route (see Appendix~\ref{appendix:VRPTW} for details).

\begin{table*}[h]
\setstretch{1.5}
    \centering
    \vspace{-0.0cm}
    \caption{ \small VRPTW: The objective values of integer solution for all small-size R2 instances (in parentheses is the solving time in seconds). The last column represents the ranking of RLHH among all the methods.}
    \label{tab:vrptw_r2_detail}
    \vspace{0.2cm}
    \scalebox{0.75}{
    \begin{tabular}{cccccccccc}
    \toprule
    No. & instance & n  & BE1           & BE2              & BE3     & BN              & BP       & \textbf{RLHH(Ours)} & Rank  \\
    \midrule
    1   & r203  & 25 & 669.77           (12)    & 669.04           (58)    & 768.30   (67)   & 689.00          (79)    & 714.12  (256)    & \textbf{520.25}  (46)   & 1 \\
    2   & r202  & 30 & \textbf{440.98}  (13)    & 522.45           (65)    & 509.47   (14)   & 516.33          (45)    & 639.64  (62)     & 480.86           (28)   & 2 \\
    3   & r201  & 27 & 520.28           (5)     & 506.05           (5)     & 506.71   (7)    & 506.05          (8)     & 507.69  (11)     & \textbf{504.17}  (7)    & 1 \\
    4   & r209  & 31 & \textbf{715.36}  (12)    & 788.04           (22)    & 858.35   (81)   & 820.94          (32)    & 1160.49 (132)    & 819.72           (7)    & 3 \\
    5   & r208  & 28 & 1291.17          (600)   & \textbf{1113.49} (540)   & 1291.17  (590)  & 1291.17         (600)   & 1232.41 (600)    & 1200.78          (4)    & 2 \\
    6   & r204  & 32 & 1504.75          (600)   & 1221.90          (31)    & 1504.75  (600)  & 1504.75         (600)   & 1408.62 (600)    & \textbf{1139.31} (8)    & 1 \\
    7   & r202  & 25 & 559.35           (6)     & 499.75           (11)    & 550.35   (12)   & 543.28          (22)    & 552.21  (29)     & \textbf{496.58}  (13)   & 1 \\
    8   & r211  & 34 & \textbf{1151.35} (161)   & 1210.89          (70)    & 1626.38  (600)  & 1352.74         (326)   & 1488.33 (600)    & 1467.85          (7)    & 3 \\
    9   & r209  & 25 & 821.21           (5)     & 739.32           (16)    & 1121.88  (47)   & 917.29          (12)    & 868.09  (63)     & \textbf{643.02}  (7)    & 1 \\
    10  & r202  & 27 & 490.71           (9)     & \textbf{463.77}  (22)    & 577.41   (9)    & 549.26          (34)    & 684.89  (53)     & 533.66           (14)   & 3 \\
    11  & r210  & 28 & 786.53           (11)    & 660.88           (34)    & 903.06   (17)   & 847.16          (40)    & 775.87  (75)     & \textbf{657.92}  (5)    & 1 \\
    12  & r211  & 26 & 1018.51          (92)    & 1085.38          (158)   & 1268.52  (600)  & 1253.23         (129)   & 1148.47 (242)    & \textbf{1007.23} (2)    & 1 \\
    13  & r202  & 28 & 534.73           (10)    & 543.75           (46)    & 537.56   (13)   & 468.72          (53)    & 526.55  (53)     & \textbf{456.28}  (30)   & 1 \\
    14  & r209  & 26 & \textbf{627.79}  (7)     & 766.72           (10)    & 723.77   (50)   & 760.53          (14)    & 747.84  (76)     & 704.66           (9)    & 2 \\
    15  & r201  & 32 & 583.40           (7)     & 545.18           (12)    & 573.84   (8)    & \textbf{545.12} (15)    & 564.58  (23)     & 561.28           (11)   & 3 \\
    16  & r203  & 29 & 876.79           (34)    & 817.75           (43)    & 792.64   (64)   & 876.15          (42)    & 1267.10 (78)     & \textbf{608.75}  (45)   & 1 \\
    17  & r205  & 33 & 912.83           (342)   & \textbf{685.97}  (199)   & 1045.75  (239)  & 957.57          (440)   & 1055.67 (289)    & 839.28           (59)   & 2 \\
    \bottomrule
    \end{tabular}}
\end{table*}

\begin{figure}[htbp]
    \centering
    \includegraphics[width=1.0\textwidth]{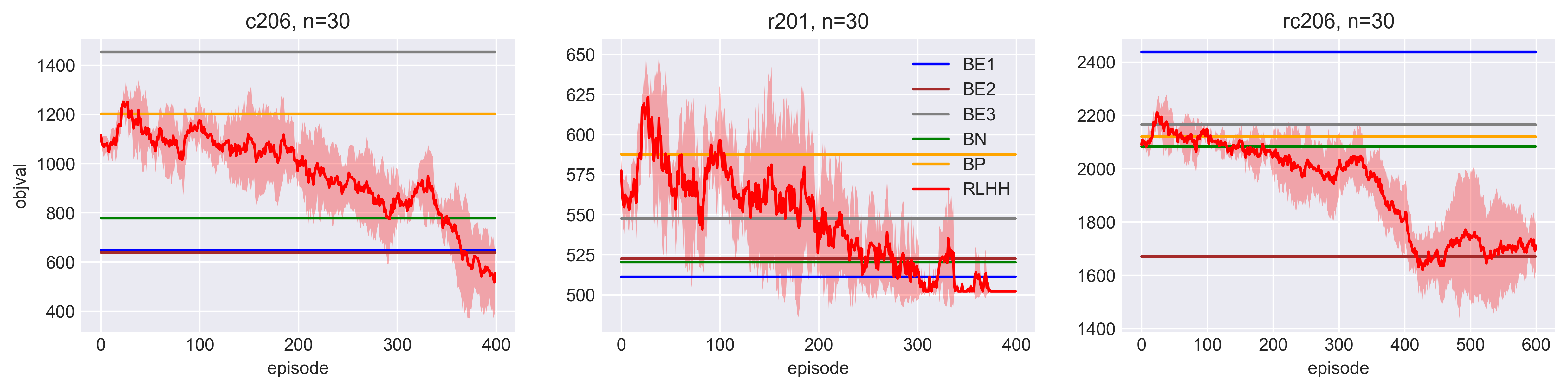}
    \vspace{-0.6cm}
    \caption{VRPTW: training curves of RL agent for fixed instances}
    \label{fig:train_episode}
    \vspace{-0.4cm}
\end{figure}

\paragraph{Datasets.}
We use the well-known Solomon benchmark \cite{solomon1987} for training and testing data.
This dataset contains six problem types (C1, C2, R1, R2, RC1, RC2), and each type has 8–12 instances with 100 customers.
``C'' refers to geographically clustered customers, ``R'' refers to randomly placed customers, and ``RC'' refers to a mixture. 
The category ``1'' and ``2'' represent narrow time windows/small vehicle capacity and large time windows/large vehicle capacity, respectively. They represent different problem structures, where the category ``2'' is more challenging.
% There are 56 instances in Solomon’s dataset, and 
From each of the original Solomon instances, we can generate smaller instances by considering only the first $n<100$ customers.

We use small-size instances from types C1, R1 and RC1 for training.
On each episode, we choose one of all instances of the three types with equal probability and consider the first $n$ customers, where $n$ is uniformly sampled from 25 to 35.
To test the generalization ability of our model, we consider two sets of instances:
(1) small-size ($25\le n \le 35$) instances from types C2, R2 and RC2;
(2) large-size ($n=50,75,100$) instances from all six types.

\paragraph{Training Process.}
We fix a problem instance to observe the performance of the RL agent in the training process. 
That is, in every episode, RL is solving the same VRPTW instances. 
We know that CG is an exact method for solving LP, so no matter what heuristic is used, the algorithm will eventually converge to the same relaxation objective value. However, since different heuristics produce different feasible routes in each CG iteration, the final integer solution may vary.
At the end of each episode, we calculate the integer solution value of the original problem. As shown in Figure~\ref{fig:train_episode}, we draw the training curves of RL agent for fixed instances. The red solid lines representing the average value using different random seeds, and the shaded region surrounding each line represents the range within one standard deviation above and below the mean. The results indicate that the RL agent does not perform as well as most low-level heuristic methods when approaching a random selection strategy in the early stage. But over the course of training, the RL agent gradually learns how to select the appropriate low-level heuristic in each iteration and can exceed all baselines after about 400 episodes. 
% The learning curve fluctuates to a certain extent, but the fluctuation amplitude gradually decreases.

\begin{sidewaystable}
\setstretch{1.5}
    \centering
    \caption{VRPTW: The average objective values for large-size instances (in parentheses is the solving time in minutes). 
    Speedup is the ratio between the solving time of the best baseline and our method.}
    \label{tab:vrptw_average_objval}
    \vspace{0.1cm}
    \scalebox{1.0}{
    \begin{tabular}{cccccccccc}
    \toprule
    Type & n   & BE1              & BE2              & BE3     & BN               & BP      & \textbf{RLHH(Ours)}    & Gain (\%)        & Speedup       \\    
    \midrule
    C1   & 50  & 479.59           (2.3)    & 487.16           (1.4)    & 484.88  (3.1)    & \textbf{447.51} (2.6)    & 482.20  (3.1)    & 461.80           (2.3)    &  -3.19    & \textbf{1.1$\times$}  \\
    C1   & 75  & 811.97           (18.8)   & 776.63           (11.5)   & 781.43  (26.8)   & 816.62          (26.1)   & 801.36  (25.2)   & \textbf{776.60}  (23.0)   &  0.00     & 0.5$\times$           \\
    C1   & 100 & 949.65           (29.1)   & 924.07           (23.9)   & 962.20  (43.7)   & 1017.13         (36.1)   & 1337.58 (37.7)   & \textbf{919.42}  (35.9)   &  0.50     & 0.7$\times$           \\    \midrule
    % C2   & 30  & 632.72           (2.1)    & 563.12           (0.6)    & 821.04  (1.8)    & 694.98          (1.6)    & 706.03  (2.4)    & \textbf{551.39}  (0.7)    &  2.08     & 0.9$\times$           \\
    C2   & 50  & 1461.24          (15.2)   & \textbf{1080.63} (3.5)    & 1461.22 (12.1)   & 1293.50         (19.2)   & 1379.02 (19.9)   & 1118.97          (2.1)    &  -3.55    & \textbf{1.6$\times$}  \\
    C2   & 75  & 2823.63          (31.9)   & 2310.73          (18.1)   & 2842.39 (24.3)   & 2879.89         (33.9)   & 2649.13 (43.0)   & \textbf{1666.66} (13.5)   &  27.87    & \textbf{1.3$\times$}  \\
    C2   & 100 & 3462.20          (37.4)   & 2828.73          (28.8)   & 3325.89 (38.2)   & 4062.15         (52.6)   & 4108.05 (56.0)   & \textbf{2598.43} (31.9)   &  8.14     & 0.9 $\times$          \\     \midrule
    R1   & 50  & 916.89           (0.7)    & 918.65           (0.6)    & 923.25  (0.9)    & 917.47          (1.4)    & 941.99  (2.0)    & \textbf{900.45}  (0.6)    &  1.79     & \textbf{1.1$\times$}  \\
    R1   & 75  & \textbf{1244.93} (4.5)    & 1260.64          (3.0)    & 1291.90 (5.3)    & 1254.84         (7.8)    & 1300.88 (11.6)   & 1253.08          (3.1)    &  -0.65    & \textbf{1.4$\times$}  \\
    R1   & 100 & \textbf{1520.71} (20.8)   & 1569.93          (9.1)    & 1559.86 (21.5)   & 1610.02         (29.1)   & 1812.55 (35.0)   & 1552.53          (10.6)   &  -2.09    & \textbf{2.0$\times$}  \\     \midrule
    % R2   & 30  & 794.44           (1.7)    & 755.31           (1.2)    & 891.76  (3.0)    & 847.02          (2.2)    & 902.50  (3.2)    & \textbf{743.62}  (0.2)    &  1.55     & \textbf{5.6$\times$}  \\
    R2   & 50  & 2048.09          (40.8)   & 1638.07          (14.9)   & 2289.24 (45.8)   & 2078.57         (33.2)   & 2300.92 (50.0)   & \textbf{1632.37} (5.6)    &  0.35     & \textbf{2.7$\times$}  \\
    R2   & 75  & 3566.78          (53.3)   & 2828.35          (32.3)   & 3600.78 (54.3)   & 3538.96         (52.0)   & 3525.12 (55.5)   & \textbf{2818.15} (33.5)   &  0.36     & 1.0$\times$           \\
    R2   & 100 & 4678.47          (55.8)   & 4223.64          (46.8)   & 4633.36 (55.6)   & 4689.24         (56.6)   & 4480.23 (58.6)   & \textbf{4184.69} (47.2)   &  0.92     & 1.0$\times$           \\     \midrule
    RC1  & 50  & 1421.91          (1.3)    & \textbf{1134.31} (0.8)    & 1429.31 (1.4)    & 1271.90         (1.7)    & 1410.54 (2.5)    & 1165.87          (0.8)    &  -2.78    & \textbf{1.0$\times$}  \\
    RC1  & 75  & 1505.07          (7.3)    & 1643.00          (3.8)    & 1635.30 (5.7)    & 1654.54         (9.4)    & 1644.06 (12.7)   & \textbf{1500.41} (2.7)    &  0.31     & \textbf{2.7$\times$}  \\
    RC1  & 100 & 1922.00          (16.1)   & \textbf{1882.32} (8.4)    & 1971.17 (14.4)   & 1968.17         (24.3)   & 2258.36 (30.1)   & 1886.23          (8.3)    &  -0.21    & \textbf{1.0$\times$}  \\     \midrule
    % RC2  & 30  & 2010.21          (3.9)    & \textbf{1746.74} (2.0)    & 2052.24 (4.1)    & 2032.23         (3.3)    & 1890.62 (4.8)    & 1819.27          (0.2)    &  -4.15    & \textbf{8.2$\times$}  \\
    RC2  & 50  & 3464.67          (31.1)   & 3288.29          (23.0)   & 3579.94 (33.2)   & 3614.39         (30.8)   & 3392.25 (38.7)   & \textbf{3134.03} (2.0)    &  4.69     & \textbf{11.4$\times$}  \\
    RC2  & 75  & 4350.91          (38.6)   & 4245.13          (28.1)   & 4514.18 (39.8)   & 4535.39         (37.2)   & 4425.03 (51.3)   & \textbf{4173.75} (21.0)   &  1.68     & \textbf{1.3$\times$}  \\
    RC2  & 100 & 5867.65          (52.0)   & 5302.88          (36.4)   & 5805.75 (51.6)   & 5864.54         (53.8)   & 5934.63 (55.7)   & \textbf{5297.46} (38.1)   &  0.10     & 1.0$\times$           \\
    \bottomrule
    \end{tabular}}
    \vspace{-0.2cm}
\end{sidewaystable}

\paragraph{Test Results.}
We evaluate the performance of the different methods using 50 small-size test instances sampled from types C2, R2, and RC2. The detail results of type R2 are shown in Table~\ref{tab:vrptw_r2_detail}. For results of other types, please refer Table~\ref{tab:vrptw_c2rc2_detail} in Appendix~\ref{appendix:VRPTW_experiments}. For most instances, the RLHH method obtains the best integer solution and remains in the top 3 for all instances.
The inclusion of computational time within parentheses significantly reduces, and all instances are solved within 1 minute by RLHH, while other methods may still remain unresolved after 10 minutes (the given time limit in small-size instances).
Although RL's agent only used C1, R1 and RC1 training data, it far outperforms other methods in C2, R2, RC2 instances, achieving the best results in most instances, which shows that RLHH has good generalization for problem structure.

In addition, we generalize the trained model on small-size instances from types C1, R1 and RC1 to large-size ($n\in\{50,75,100\}$) instances from all types. We set the total solving time limit to 1 hour in this part of experiments. For all instances of a fixed type and size, we calculate the average of their objective values and report the mean computation time in minutes (see Table~\ref{tab:vrptw_average_objval}). The last two columns show the reduction of the objective value gained and the speedup in comparison with the best-performing baseline heuristic method in each case. For type C2 with $n=75$, the RLHH obtains the reduction in the objective value of up to 27.87\% within only 74.6\% computation time (i.e. 1.34 times faster).  

According to the results, our approach significantly improves the best low-level heuristic in most cases. It shows that our method successfully combines the strength of OR heuristics with the generalization capabilities of RL, learning how to select the appropriate low-level heuristic in each CG iteration. Furthermore, we test some larger instances with $n\ge 100$ to show the generalization capability of our method in problem size (see Table~\ref{tab:vrptw_large} in Appendix~\ref{appendix:VRPTW_experiments}).

\begin{figure*}
    \centering
    \includegraphics[width=1.0\textwidth]{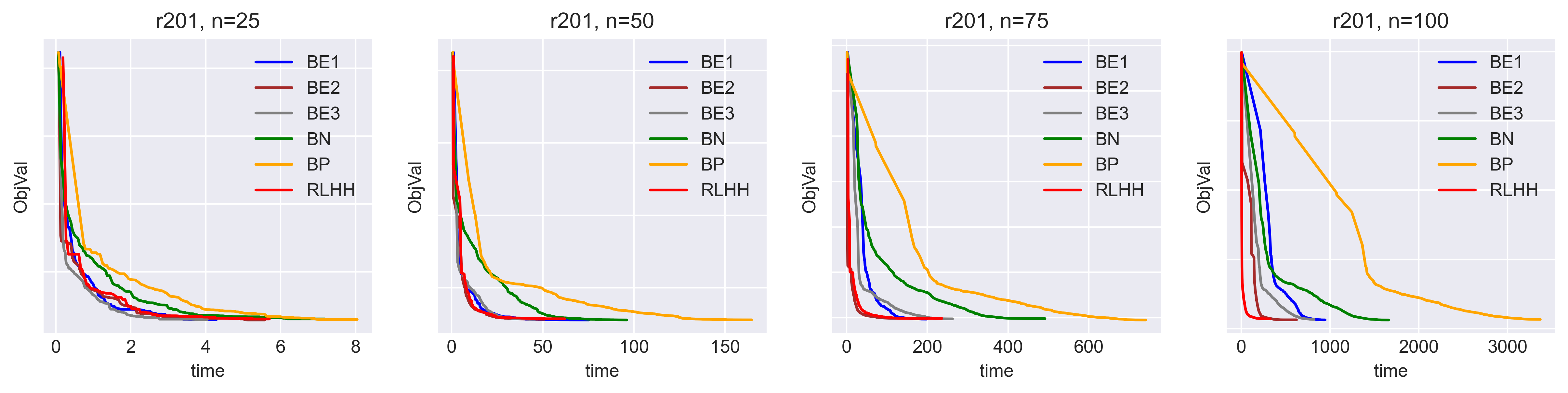}
    \vspace{-0.6cm}
    \caption{VRPTW: Iteration process of all methods under different size (time is in seconds)}
    \label{fig:iteration_process_vrptw}
    \vspace{-0.2cm}
\end{figure*}

\paragraph{Convergence Process.}
We record the objective values of the RMP at each CG iteration for the given method using $r201$ as a test instance, considering instance size from 25 to 100 (see Figure~\ref{fig:iteration_process_vrptw}).
% considering the first 25, 50, 75, and all 100 customers 
There is little difference between the methods when $n\le 50$, with BP and BN performing slightly worse. However, when the number of customers grows large, the advantages of RLHH become evident. In the case of $n=100$, our method surpasses all others and only converges about 1/2 CPU time of the best low-level heuristic BE2. RLHH dominates almost all other methods throughout the CG iterations in large-size situations. It means that if CG had to be terminated early, RLHH would produce the best (minimum) objective value compared to other methods.

\subsection{Bus Driver Scheduling Problem}
\label{section:BDSP}

BDSP, a kind of important problem in the field of public transport, aims to develop a set of schedules (each schedule requires a bus driver) that cover all timetabled bus trips and abide by various labor union regulations. A feasible driver schedule consists of trips with enough free time between trips for drivers to change shifts and total working hours that meet labor law requirements. The goal of BDSP is to use the fewest drivers to complete a certain number of trips and minimize the total working time of all drivers. We can model the problem as MP in Section~\ref{section:CG}, and each column corresponds to a feasible schedule (see Appendix~\ref{appendix:BDSP} for details).

\begin{figure*}
    \centering
    \includegraphics[width=1.0\textwidth]{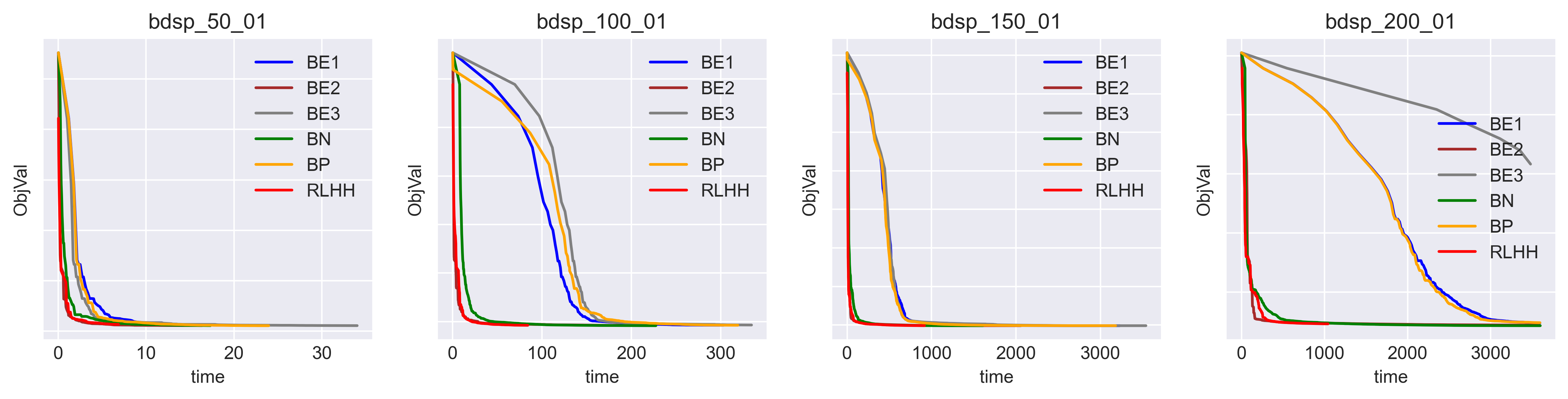}
    \vspace{-0.6cm}
    \caption{BDSP: Iteration process of all methods under different size instances (time is in seconds)}
    \label{fig:iteration_process_bdsp}
    \vspace{-0.2cm}
\end{figure*}

\paragraph{Datasets.}
As far as we know, there is no public data set for BDSP, so we construct a random data generator referring \cite{Haase2001vcsp}. By taking a total number of trips as an input, the trips are randomly distributed over time, with more trips during peak hours. Specifically, the starting hour of each trip is determined according to the probability distribution given in Table~\ref{tab:start_hour} of Appendix~\ref{appendix:BDSP}, and the starting minute is uniformly distributed. Then we randomly sample an integer between 60 and 90 as the duration (in minutes) for each trip.
We generate instances with $50\le n\le 60$ for training, and $30\times 5=150$ new instances with $n\in \{50, 75, 100, 150, 200\}$ are generated for testing.
We set the solution time limit to 10 minutes for instances with $n\le 100$ and 1 hour for larger-size instances.

\paragraph{Test Results.}
As before, Table~\ref{tab:bdsp_average_objval} shows the average objective value for different size instances. It illustrates that the RLHH performs better on large-size problem instances. When $n\ge 100$, RLHH exceeds the other methods, and the objective value at $n=200$ is 15.41\% lower than the best low-level heuristic, which is a great improvement. And in average, the RLHH is faster than the best baseline for instances of all sizes (when $n=200$, none of the methods can complete CG process in limited time).

\begin{table}[!htbp]
\setstretch{1.5}
    \centering
    \vspace{-0.0cm}
    \caption{BDSP: The average objective values for test instances (in parentheses is the solving time in minutes). Speedup is the ratio between the solving time of the best baseline and our method.}
    \label{tab:bdsp_average_objval}
    \scalebox{0.8}{
    \begin{tabular}{cccccccccc}
    \toprule
    n   & BE1     & BE2             & BE3     & BN              & BP      & \textbf{RLHH(Ours)}    & Gain (\%)     & Speedup\\
    \midrule
    50  & 6736  (1.4)   & 5644          (0.7)   & 5972  (0.6)   & \textbf{5605} (0.6)   & 6542  (0.8)  & 5627           (0.2)   & -0.39   & \textbf{2.3}$\times$  \\
    75  & 12829 (6.8)   & \textbf{7538} (3.6)   & 11307 (5.2)   & 10039         (4.1)   & 11389 (5.4)  & 7995           (1.9)   & -6.06   & \textbf{1.9}$\times$  \\
    100 & 15252 (10.4)  & 10796         (9.9)   & 14837 (10.3)  & 11958         (10.8)  & 18097 (10.1) & \textbf{9933}  (8.2)   & 7.99    & \textbf{1.2}$\times$  \\
    150 & 18887 (57.7)  & 14597         (60.0)  & 20926 (60.0)  & 19005         (60.0)  & 17681 (58.3) & \textbf{13552} (57.9)  & 7.16    & \textbf{1.0}$\times$  \\
    200 & 26303 (60.0)  & 27858         (60.0)  & 42671 (60.0)  & 22762         (60.0)  & 25441 (60.0) & \textbf{19255} (60.0)  & 15.41   & \textbf{1.0}$\times$  \\
    \bottomrule
    \end{tabular}}
\end{table}

\paragraph{Convergence process.}
The CG process of different methods, considering instance size from 50 to 200, is shown in Figure~\ref{fig:iteration_process_bdsp}. The names of the instances are written in the form ``bdsp\_\textit{Size}\_\textit{Id}'', where \textit{Size} is the number of trips, and \textit{Id} is the instance identifier. Similar to BE2 and BN, RLHH converges quickly and is significantly better than other methods.

\section{Conclusion}
In this study, we provide a way of combining the strength of OR heuristics with the learning and generalization capabilities of RL models. Specifically, we propose an RL-based hyper-heuristic enhancing the CG method for vehicle routing and scheduling problems, where the PP is a SPPRC or its variants defined on a network. At each CG iteration, the RL agent selects a low-level heuristic to generate a reduced network that keeps only the most promising edges. The comparative experiments show that the proposed algorithm can combine the advantage of different low-level heuristics after appropriate training to achieve significant improvements compared with the single heuristic algorithm.
Moreover, our RL model demonstrates strong generalization capabilities, exhibiting the capacity to be trained on small-scale instances while effectively extrapolating its learned policies to larger and more complex scenarios.
% The RL model is trained on the small-size data by the DDQN algorithm with experience replay and can be well generalized to large-size instances.

% \paragraph{Future Work.}
The reinforcement learning task considered in this paper is relatively straightforward. In the future, it may be worthwhile to enable the agent to select low-level heuristics while simultaneously determining their corresponding parameters. This could potentially necessitate a more detailed exploration of the state space, perhaps employing techniques such as graph neural networks. Alternatively, we can use policy-based RL methods to tackle more complex decision spaces. Furthermore, our approach can be extended to more difficult scheduling problems, such as railway crew scheduling problems and airline crew scheduling problems, as well as other graph-based CO problems suitable for the CG framework, such as graph coloring problems.

% \textbf{Limitation.}
% Our method is not yet able to obtain the exact solution, so it cannot be compared with the VRPTW results published on the SINTEF website \cite{vrptw-web}. We follow \cite{morabit2021} to solve the root node of the B\&P search tree, while the proposed RLHH strategy is also applicable to other nodes. The experiments show that our method can obtain better integer solutions (upper bounds) in less time, which implies that it can efficiently get exact solutions in B\&P.

\newpage

\bibliographystyle{apalike}
\bibliography{references}

\appendix
\onecolumn
In this appendix, we provide the details to help readers better understand our proposed approach. 
In Appendix~\ref{appendix:RLHH}, we give the detailed training procedure in Algorithm~\ref{alg:RLHH_train}. 
In Appendix~\ref{appendix:VRPTW} and Appendix~\ref{appendix:BDSP}, we provide the problem definition and modeling of VRPTW and BDSP, respectively. Finally, in Appendix~\ref{appendix:VRPTW_experiments}, we present additional experimental results for VRPTW.

\section{Training algorithm of RLHH}
\label{appendix:RLHH}

Deep Q-network (DQN) is a widely used RL method, which combines a deep neural network function approximator with the classical Q-learning algorithm to learn a state-action value function $Q(s,a)$. The network outputs a vector of actions' (i.e. low-level heuristics) scores as $Q(s,\cdot;\theta)$, where $\theta$ is the parameters of the deep neural network.
In DQN, we want the output $Q(s_i,a;\theta)$ of the network for each state $s_i$ to be as close to the Temporal Difference (TD) target $y_i$ as possible:
\begin{equation}
    y_i = r_i + \gamma \max_a Q(s_i',a;\theta).
    \label{y_dqn}
\end{equation}

Due to the max operator in standard DQN (see \eqref{y_dqn}), DQN agents are biased to select overestimated values, resulting in overly optimistic value estimates. Therefore \cite{van2016deep} introduce Double-DQN (DDQN) to reduce overestimation by decomposing the max operation into two steps: first select the optimal action in state $s$, then calculate the value corresponding to the action. In the second step, another network named target network (with parameters $\theta_t^-$) is used. The two networks share the same structure, but the parameters of the target network are only updated every $\tau$ steps or episodes from the online network. The target used by DDQN is defined as:
\begin{equation}
    y_i = r_i + \gamma Q(s_i',\arg\max_a Q(s_i',a;\theta);\theta^-).
    \label{y_ddqn}
\end{equation}
In addition, the loss function used in the training process is defined as:
\begin{equation}
    L(\theta) = \sum_i(y_i-Q(s_i,a;\theta))^2.
\end{equation}
A multilayer perceptron is used as Q-function approximator in our experiments.

Algorithm~\ref{alg:RLHH_train} describes the details of the RLHH training process.
\begin{algorithm}
\setstretch{1.5}
    \caption{Training procedure of RLHH: RL-based hyper-heuristic in CG}
    \label{alg:RLHH_train}
    \begin{algorithmic}[1] %[1] enables line numbers
        \STATE Initialize parameters of both the target network and online network with random values $\theta$
        \STATE Initialize experience replay buffer $M$ with capacity $L$
        \FOR{each episode}
            \STATE Let $t=0$; Choose a training instance.
            \STATE Build the RMP$_0$ with initial columns of the problem instance.
            \STATE Solve RMP$_0$; Use dual values to get complete pricing network.
            \STATE Calculate initial state $s_0$.
            \WHILE{exist new columns with negative reduced cost}
            
            \STATE Choose action $a_t$ (a low-level heuristic) according to $\epsilon$-greedy policy.
            \STATE Use selected low-level heuristic to get a reduced network.
            \STATE Attempt to find a new negative reduced cost column in the reduced network. 
            \IF{not found}
            \STATE Find a new negative reduced cost column in the complete network.
            \ENDIF
            \STATE Calculate the reward of the current step $r_t$
            \STATE Add new column to RMP$_t$ and obtain RMP$_{t+1}$;
            \STATE Solve RMP$_t$; Use dual values to get the complete pricing network.
            \STATE Calculate next state $s'_{t}$.
            \STATE Store transition $(s_t,a_t,r_t,s'_{t})$ in the buffer $M$.
            \STATE Set $t = t+1$.
            \STATE Set $b \leftarrow$ A randomly sampled mini-batch of transitions from the buffer $M$.
            \FOR{each transition $(s_i,a_i,r_i,s_i')$ in $b$}
                \IF {$s_i'$ is a terminal state}
                \STATE $y_i = r_i$;
                \ELSE
                \STATE $y_i = r_i + \gamma Q(s_i',\arg\max_a Q(s_i',a;\theta);\theta^-)$
                \ENDIF
            \ENDFOR
            \STATE Calculate the loss function $L(\theta) = \sum_i(y_i-Q(s_i,a;\theta))^2$, use gradient descent to update parameters $\theta$.
            \STATE Update the parameters of the target network: set $\theta^- \leftarrow \theta$ every $\tau$ steps.
            \ENDWHILE
            % \STATE Set the variable to integer type and solve MIP.
        \ENDFOR
    \end{algorithmic}
\end{algorithm}

% \newpage

\section{Problem Setup of VRPTW}
\label{appendix:VRPTW}
VRPTW aims to find feasible routes (each route requires a vehicle) to deliver goods to a geographically dispersed group of customers while minimizing total travel costs. 
Given a set of customers where each customer is specified a priori by demand, a position and a time window within which the customer must be served. Our problem is to determine the number of vehicles required, together with their routes, so that each customer is covered while the fixed costs related to vehicles, and the travel costs between customers, are minimized. The vehicle must begin service within the customer's time window. The load capacity per vehicle is limited, and the travel time between two customers is proportional to the Euclidean distance between their positions.

Let $G=(V,E)$ be a directed graph with a set of nodes $V$ that include the source and destination nodes representing the depot denoted by $s$ and $t$, respectively, and $E$ the set of edges. The nodes represent the customers, and the edges represent the compatibility of two customers with respect to time and demand. There exist a directed edge $(i,j)\in E$ if 
\begin{equation}
    \overline{w}_j - \underline{w}_i \ge T_i + t_{ij},
\end{equation}
where $\overline{w}_j$ and $\underline{w}_i$ represent the latest service start time of customer $j$ and the earliest service start time of customer $i$ respectively, $t_{ij}$ is the travel time from node $i$ to node $j$, and $T_i$ is the service time for customer node $i$. The source node $s$ can go to any customer node $i\in A\backslash \{s,t\}$, and any customer node can go to the sink node $t$.

A feasible path from the source $s$ to the sink $t$ in the network represents a feasible route of a vehicle with some goods originating and terminating at the depot for a fleet of vehicles which services a set of customers with known demands. We define the cost of each edge $(i,j)\in E$ as the Euclidean distance between two nodes denoted $c_{ij}$. By setting the fixed cost of the vehicle on edges from the source node $s$, the problem of finding new routes (i.e., the pricing problem) is an ESPPRC, a variant of SPPRC, which requires all paths is elementary (each node visit at most once) with modified edge costs described in Section~\ref{appendix:SPPRC}.

% \newpage
\section{Problem Setup of BDSP}
\label{appendix:BDSP}

BDSP aims to find feasible schedules (each schedule requires a bus driver) that cover all timetabled bus trips and abide by various labor union regulations while minimizing total operational costs.
Consider a set of bus trips where each trip is specified a priori by a starting time, a duration and a cost. Our problem is determining the number of drivers required, together with their schedules, so each trip is covered while the fixed costs related to drivers and the travel costs between trips are minimized. There are limits on how many hours each driver can drive, work and drive continuously (driving time counts only for time on the bus, while working time includes breaks between trips).

Let $G=(V,A)$ be a directed graph with a set of nodes $V$ that include the virtual source and destination nodes denoted by $s$ and $t$, respectively, and $E$ the set of edges. The nodes represent the timetabled bus trips, and the edges represent the time compatibility of the two trips. There exist a directed edge $(i,j)\in E$ if there is enough free time between the trips $i,j\in V$, it is easy to know that the start time of $j$ must be after the end time of $i$, which is different from VRPTW. That is:
\begin{equation}
    \underline{w}_j - \overline{w}_i \ge T,
\end{equation}
where $\underline{w}_j$ and $\overline{w}_i$ represent the start time of trip $j$ and the end time of trip $i$ respectively, and $T$ is the minimum time difference for drivers between trips.

A feasible path from the source $s$ to the sink $t$ in the network represents a feasible schedule for a driver during the day. Apparently, the underlying network is acyclic, and all paths are elementary from the definition.
We define the cost of each edge $(i,j)\in E$ as the driving time of trip $j$ plus the break time between two trips:
\begin{equation}
    c_{ij} = (\overline{w}_j - \underline{w}_j) + (\underline{w}_j - \overline{w}_i) = \overline{w}_j - \overline{w}_i.
\end{equation}
By setting the fixed cost of the vehicle on edges from the source node $s$, the problem of finding new schedules (i.e., the pricing problem) is an SPPRC.

\begin{table}[htbp]
\setstretch{1.5}
    \caption{BDSP: Trip starting hour probability distribution}
    \label{tab:start_hour}
    \centering
    \begin{tabular}{ccccccccccccc}
        \toprule
        Hours (a.m.) & 0 & 1 & 2 & 3 & 4 & 5 & 6 & 7 & 8 & 9 & 10 & 11 \\
        Prob. (\%) & 0 & 0 & 0 & 0 & 3 & 3 & 5 & 9 & 10 & 8 & 5 & 4 \\
        \midrule
        Hours (p.m.) &  0 & 1 & 2 & 3 & 4 & 5 & 6 & 7 & 8 & 9 & 10 & 11 \\
        Prob. (\%) & 3 & 3 & 4 & 5 & 9 & 10 & 8 & 5 & 3 & 3 & 0 & 0 \\
        \bottomrule
    \end{tabular}
\end{table}

\begin{table}
\setstretch{1.5}
    \centering
    % \vspace{-0.35cm}
    \caption{VRPTW: The objective values of integer solution for all small-size C2 and RC2 instances (in parentheses is the solving time in seconds). Rank represents the ranking of RLHH solution. Speedup is the ratio between the solving time of the best baseline and our method.}
    \label{tab:vrptw_c2rc2_detail}
    \scalebox{0.7}{
    \begin{tabular}{ccccccccccc}
        \toprule
    No. & instance & n  & BE1         & BE2            & BE3            & BN           & BP           & \textbf{RLHH(Ours)} & Rank  & Speedup \\
    \midrule
    1   & c208  & 28 & 1450.31 (11)   & 993.39  (6)    & 1391.35 (7)    & 1408.24 (15)   & 1261.36 (34)   & \textbf{883.14}  (11)  & 1 & 0.35$\times$ \\
    2   & c202  & 31 & 476.01  (93)   & 404.60  (37)   & 300.69  (134)  & 283.09  (68)   & \textbf{257.53}  (142)  & 404.60  (55)  & 4 & \textbf{2.56$\times$} \\
    3   & c201  & 30 & \textbf{233.40}  (25)   & \textbf{233.40}  (42)   & \textbf{233.40}  (13)   & \textbf{233.40}  (33)   & \textbf{233.40}  (26)   & \textbf{233.40}  (12)  & 1 & \textbf{1.09$\times$} \\
    4   & c202  & 29 & \textbf{227.76}  (34)   & 386.56  (36)   & 279.81  (66)   & 334.69  (83)   & 227.76  (96)   & 386.56  (43)  & 5 & 0.81$\times$ \\
    5   & c201  & 33 & \textbf{250.30}  (60)   & \textbf{250.30}  (60)   & \textbf{250.30}  (27)   & \textbf{250.30}  (65)   & \textbf{250.30}  (53)   & \textbf{250.30}  (27)  & 1 & 0.99$\times$ \\
    6   & c202  & 34 & \textbf{250.38}  (600)  & 409.18  (74)   & 302.43  (93)   & 443.99  (264)  & 250.38  (338)  & 409.18  (73)  & 4 & \textbf{8.26$\times$} \\
    7   & c202  & 26 & \textbf{220.21}  (61)   & 328.67  (35)   & 272.26  (25)   & \textbf{220.21}  (43)   & \textbf{220.21}  (134)  & \textbf{220.21}  (25)  & 1 & \textbf{1.02$\times$} \\
    8   & c206  & 32 & \textbf{717.58}  (13)   & 843.10  (13)   & 1636.60 (9)    & 1448.43 (27)   & 1392.82 (33)   & 872.60  (20)  & 3 & 0.66$\times$ \\
    9   & c204  & 34 & 1793.70 (600)  & 1135.74 (148)  & 1793.70 (600)  & 1793.70 (600)  & 1677.51 (600)  & \textbf{1058.21} (84)  & 1 & \textbf{1.75$\times$} \\
    10  & c201  & 25 & 215.54  (6)    & 215.54  (17)   & 215.54  (7)    & 215.54  (16)   & 215.54  (12)   & \textbf{215.54}  (6)   & 1 & \textbf{1.03$\times$} \\
    11  & c206  & 30 & 648.13  (12)   & \textbf{639.15}  (12)   & 1453.26 (6)    & 778.28  (24)   & 1201.99 (24)   & 882.51  (14)  & 4 & 0.83$\times$ \\
    12  & c208  & 25 & 1080.12 (9)    & 851.73  (6)    & 1141.49 (4)    & 1144.98 (8)    & 1018.48 (25)   & \textbf{798.41}  (7)   & 1 & 0.91$\times$ \\
    13  & c208  & 34 & 1626.33 (42)   & 1140.98 (11)   & 1793.70 (68)   & 1664.45 (28)   & 1549.63 (79)   & \textbf{1128.33} (16)  & 1 & 0.69$\times$ \\
    14  & c203  & 31 & 460.16  (337)  & 545.21  (42)   & 1598.63 (600)  & \textbf{426.89}  (158)  & 1014.05 (600)  & 446.97  (146) & 2 & \textbf{1.08$\times$} \\
    15  & c202  & 27 & \textbf{223.20}  (48)   & 382.00  (25)   & \textbf{223.20}  (31)   & \textbf{223.20}  (53)   & 275.25  (76)   & \textbf{223.20}  (41)  & 1 & \textbf{1.17$\times$} \\
    16  & c207  & 31 & 1379.21 (180)  & 1123.84 (20)   & 1352.32 (33)   & 1290.01 (24)   & 1186.98 (197)  & \textbf{1092.99} (19)  & 1 & \textbf{1.06$\times$} \\
    17  & c205  & 31 & 505.01  (22)   & \textbf{465.83}  (17)   & 522.04  (8)    & 523.20  (21)   & 530.77  (27)   & 487.36  (9)   & 2 & \textbf{1.87$\times$} \\
    \midrule
    1  & rc205 & 26 & 1369.13 (7)    & 1067.94 (5)    & 1366.80 (7)    & 1283.25 (16)   & 1260.57 (18)   & \textbf{1065.13} (3)  & 1 & \textbf{1.42$\times$}  \\
    2  & rc202 & 32 & 1910.62 (26)   & 1817.91 (24)   & 1835.09 (13)   & 1937.30 (34)   & 1718.64 (107)  & \textbf{1470.31} (6)  & 1 & \textbf{3.95$\times$}  \\
    3  & rc205 & 29 & 1448.43 (8)    & \textbf{1171.07} (4)    & 1805.94 (5)    & 1549.85 (16)   & 1473.42 (19)   & 1351.56 (5)  & 2 & 0.80$\times$           \\
    4  & rc208 & 25 & 1886.66 (402)  & 1723.67 (49)   & 1886.66 (600)  & 1886.66 (135)  & 1728.48 (600)  & \textbf{1647.81} (6)  & 1 & \textbf{8.42$\times$}  \\
    5  & rc206 & 30 & 2437.55 (5)    & 1670.57 (11)   & 2165.18 (10)   & 2083.08 (15)   & 2120.83 (19)   & \textbf{1597.36} (4)  & 1 & \textbf{2.77$\times$}  \\
    6  & rc204 & 30 & 2437.55 (600)  & \textbf{2234.53} (97)   & 2437.55 (600)  & 2437.55 (600)  & 2279.37 (600)  & 2379.11 (9)  & 3 & \textbf{10.45$\times$} \\
    7  & rc204 & 34 & 2838.11 (600)  & 2838.11 (600)  & 2838.11 (600)  & 2838.11 (600)  & \textbf{2679.93} (600)  & 2779.67 (65) & 2 & \textbf{9.22$\times$}  \\
    8  & rc208 & 34 & 2838.11 (600)  & 2838.11 (600)  & 2838.11 (600)  & 2838.11 (178)  & \textbf{2679.93} (600)  & 2702.94 (48) & 2 & \textbf{12.49$\times$} \\
    9  & rc201 & 29 & 958.79  (7)    & 1110.15 (6)    & 1059.12 (3)    & 876.30  (7)    & 1066.24 (9)    & \textbf{799.62}  (5)  & 1 & \textbf{1.09$\times$}  \\
    10 & rc202 & 25 & 1150.59 (13)   & \textbf{899.59}  (13)   & 1026.36 (4)    & 1148.64 (16)   & 933.20  (93)   & 923.87  (9)  & 2 & \textbf{1.46$\times$}  \\
    11 & rc201 & 34 & \textbf{964.97}  (10)   & 1150.76 (9)    & 1296.43 (6)    & 988.77  (12)   & 1170.55 (17)   & 1013.22 (9)  & 3 & \textbf{1.03$\times$}  \\
    12 & rc206 & 33 & 2230.44 (12)   & \textbf{1490.29} (14)   & 2329.40 (17)   & 2743.24 (14)   & 2071.19 (32)   & 1809.62 (5)  & 2 & \textbf{2.95$\times$}  \\
    13 & rc203 & 33 & 2743.24 (600)  & \textbf{2311.12} (31)   & 2617.46 (540)  & 2743.24 (600)  & 2585.06 (600)  & 2359.38 (16) & 2 & \textbf{1.97$\times$}  \\
    14 & rc208 & 31 & 2538.13 (600)  & 2538.13 (233)  & 2538.13 (600)  & 2538.13 (600)  & 2379.95 (600)  & \textbf{2379.36} (19) & 1 & \textbf{12.43$\times$} \\
    15 & rc207 & 33 & 2400.89 (50)   & 2321.61 (67)   & 2743.24 (111)  & 2591.14 (165)  & 2211.96 (389)  & \textbf{2075.93} (7)  & 1 & \textbf{10.30$\times$} \\
    16 & rc207 & 30 & 1953.55 (19)   & 1600.32 (5)    & 2222.81 (33)   & 2114.00 (94)   & 1928.41 (120)  & \textbf{1600.32} (7)  & 1 & 0.71$\times$           \\
    \bottomrule
    \end{tabular}}
    % \end{adjustbox}
\end{table}

\section{More experiment results for VRPTW}
\label{appendix:VRPTW_experiments}

\paragraph{Detailed results of small-size instances.}
The detail results of type C2 and RC2 are shown in Table~\ref{tab:vrptw_c2rc2_detail}. Overall, our method obtains the best integer solution for 27 out of 50 small-size instances and remains in the top 3 for almost all instances. In addition, the computation time significantly reduces, especially for the most difficult RC2 instances.

\paragraph{Stronger experimental evaluation.}
We test some larger instances from SINTEF website \footnote{https://www.sintef.no/projectweb/top/vrptw/} to show the generalization capability in problem size of our method. Specifically, we used the first 3 instances of class C1 and C2 with $n=200, 400 , 600$ respectively, and limited the solving time to 1 hour. Almost all experiments are timeout, so we only report average costs in Table~\ref{tab:vrptw_large}, which show the scalability of our approach.

\begin{table}
\setstretch{1.5}
    \centering
    \caption{VRPTW: The average objective values for larger instances.}
    \label{tab:vrptw_large}
    % \scalebox{1.0}{
    \begin{tabular}{ccccccccc}
    \toprule
    type    & n & BE1   & BE2      & BE3        & BN        & BP    & \textbf{RLHH(Ours)} & Gain (\%) \\    
    \midrule
    C1 & 200 & 8107.1   & \textbf{5329.1}   & 8012.2   & 6987.9   & 6723.9   & 5331.9   & -0.05 \\
    C1 & 400 & 48706.4  & 58509.2  & 63920.6  & 71836.1  & 44650.1  & \textbf{37388.1}  & 16.26 \\
    C1 & 600 & 103573.3 & 95040.2  & 94502.6  & 132523.0 & 94019.5  & \textbf{92412.6}  & 1.71  \\
    \midrule
    C2 & 200 & 14355.2  & 16072.9  & 14476.8  & 14899.4  & 14290.1  & \textbf{13302.4}  & 6.91  \\
    C2 & 400 & 41170.4  & 43057.1  & \textbf{41038.1}  & 44962.2  & 45267.8  & 41478.7  & -1.07 \\
    C2 & 600 & 108828.6 & 101109.5 & 118854.0 & 115148.1 & 124204.7 & \textbf{100849.6} & 0.26  \\
    \bottomrule
    \end{tabular}
    % \end{adjustbox}
\end{table}
% \newpage

\end{document}